\documentclass{article}
\usepackage{spconf,amsmath,graphicx}
\usepackage{subcaption}
\usepackage{bm}
\usepackage{booktabs}
\usepackage{url}
\usepackage{hyperref}
\usepackage[utf8]{inputenc}
\graphicspath{{./images/}}
\usepackage{pgfplots}
\title{Image-based Survival Prediction for Lung Cancer Patients using CNNs}

\name{Christoph Haarburger\thanks{$^*$ Equal contribution}\,$^*$, Philippe Weitz\,$^*$, Oliver Rippel, Dorit Merhof}
\address{Institute of Imaging and Computer Vision, RWTH Aachen University, Germany}

\begin{document}
%
\maketitle
\begin{abstract}
	\noindent Traditional survival models such as the Cox proportional hazards model are typically based on scalar or categorical clinical features.
	With the advent of increasingly large image datasets, it has become feasible to incorporate quantitative image features into survival prediction.
	So far, this kind of analysis is mostly based on radiomics features, i.e. a fixed set of features that is mathematically defined a priori.
	To capture highly abstract information, it is desirable to learn the feature extraction using convolutional neural networks.
	However, for tomographic medical images, model training is difficult because on the one hand, only few samples of 3D image data fit into one batch at once and on the other hand, survival loss functions are essentially ordering measures that require large batch sizes.
	In this work, we show that by simplifying survival analysis to median survival classification, convolutional neural networks can be trained with small batch sizes and learn features that predict survival equally well as end-to-end hazard prediction networks.
	Our approach outperforms the previous state of the art in a publicly available lung cancer dataset.
\end{abstract}
\section{Introduction}\label{sec:introduction}

\noindent The medical image computing (MIC) community has been influenced strongly by advancements in machine learning and computer vision.
Public availability of large annotated datasets has highly improved applicability and reproducibility of deep learning in MIC.\@
As a result, the state of the art in computer aided diagnosis and detection as well as segmentation of medical images is currently dominated by convolutional neural networks (CNNs)~\cite{litjens_2017}.
A MIC subfield that has not seen such a strong benefit from these methods yet is survival prediction (i.e.~prognosis) based on medical images.
Survival analysis and prediction have been influenced mostly from biostatistics, i.e.~statistical modeling based on non-image data.
Motivated by the recent success of radiomics~\cite{aerts_2014}, there has been increasing interest in image-based survival analysis.

\subsection{Survival Analysis}

\noindent Survival analysis refers to the study of the time-to-event data for an individual or the study of the distribution of those times for a cohort.
Typical events in a medical context are death, disease incidence or relapse from remission.
Usually, regression modeling strategies cannot be applied to survival data since although for each patient, a time-to-event is specified, those events may be qualitatively different.
For some patients, the time indicated is the time-to-event, for others it indicates the time of the last follow-up before leaving the study.
This is referred to as \textit{right-censoring} and indicated by the event indicator \(\delta_i\) that equals 1 if the event occurred and 0 for censoring.

A common approach to survival analysis is the prediction of hazards \(\lambda\) from which a survival time can be obtained.
The most broadly applied model for hazard prediction is the \textit{Cox proportional hazards model}~\cite{cox_1972}, which determines patient-individual hazards \(\lambda_i\) based on covariates \(x_i\) with
\begin{equation}
\label{eqn:CPH}
\lambda(t|x)_i = \lambda_0(t) \cdot \exp\left( \beta^T x_i\right).
\end{equation}
It can ben shown that hazard prediction is essentially an ordering task~\cite{harrell_2015}.
If \(S(x_i)\) denotes the survival time of patient \(i\), two observations are correctly ordered if
\begin{equation}
\label{eqn:concordant_pair}
S(x_i)>S(x_j) \rightarrow  \lambda(x_i) < \lambda(x_j).
\end{equation}
If this holds true for the predicted hazards of two observations, they are referred to as \textit{concordant}. Correspondingly, the most broadly applied metric in survival analysis is the \textit{concordance index} or \textit{c-index}, which is defined as
\begin{equation}
\label{eqn:c_index}
C = \frac{\# \; \text{concordant pairs}}{\# \; \text{possible pairs}} \in [0,1].
\end{equation}

\noindent It can be interpreted the same way as the area under the receiver operating curve.

For several reasons, image-based survival analysis has not yet fully benefitted from recent advancements in deep neural networks:
Training data is typically censored and cannot be handled properly by classification or regression approaches.
Therefore, the most widely-used loss functions and network architectures are not applicable.
Moreover, the standard evaluation measure in survival analysis, the \textit{concordance index}, is an \textit{ordering measure} that can be hard to interpret, especially when combined with batch-wise gradient descent methods.

\subsection{Related Work and Contributions}
\noindent Most approaches to image-based survival analysis perform a large-scale image feature extraction and feature selection, followed by a linear combination of the selected features in a Cox model~\cite{aerts_2014, leger_2017, simpson_2017, attiyeh_2018}.
Recently, modern neural networks were employed for survival analysis based on non-image data in~\cite{katzman_2016, luck_2017, lee_2017a}, significantly outperforming traditional methods such as Cox models.
However, these models did not incorporate a trainable image feature extraction as needed for image-based survival prediction.
In~\cite{zhu_2016}, convolutional neural networks (CNNs) were first utilized for end-to-end trainable image feature extraction and survival analysis based on pathology images.
This model was further extended in~\cite{zhu_2017b, li_2018} to capture information from whole-slide images.
The method proposed in~\cite{yao_2017} can perform survival analysis based on both pathology images and scalar clinical data by maximizing correlation between clinical and CNN features.
To our best knowledge, there is no literature on survival analysis based on trainable image features from \textit{tomographic} images so far.
In~\cite{lao_2017}, features from a CNN trained for RGB image classification were extracted for survival prediction based on magnetic resonance images. However, in this work the CNN was not actually trained on tomographic medical images but used as a fixed feature extractor.
Tomographic medical image data is especially challenging to combine with survival prediction networks:
On one hand, due to high dimensionality, tomographic medical images require small batch sizes during training to fit into GPU memory.
On the other hand, the loss function typically used in survival analysis, the Cox partial log likelihood loss, is an ordering measure for which large sample sizes are beneficial.

We aim to address this issue by transferring features learned by a classification problem to survival analysis without losing performance.
Moreover, we propose a method to combine radiomics and learned CNN features that enforce the CNN to learn features that are both discriminative and not covered by the radiomics feature set.
All methods are evaluated on a publicly available dataset of computed tomography (CT) images of non-small-cell lung cancer (NSCLC) patients and corresponding survival labels.
We show that our method can outperform the previous state-of-the-art presented in~\cite{aerts_2014}.

\section{Methods}\label{sec:methods}
\subsection{The Lung1 Dataset}\label{sub:The Lung1 data set}
\begin{figure}[htb]
	\centering
	\includegraphics[width=0.3\textwidth]{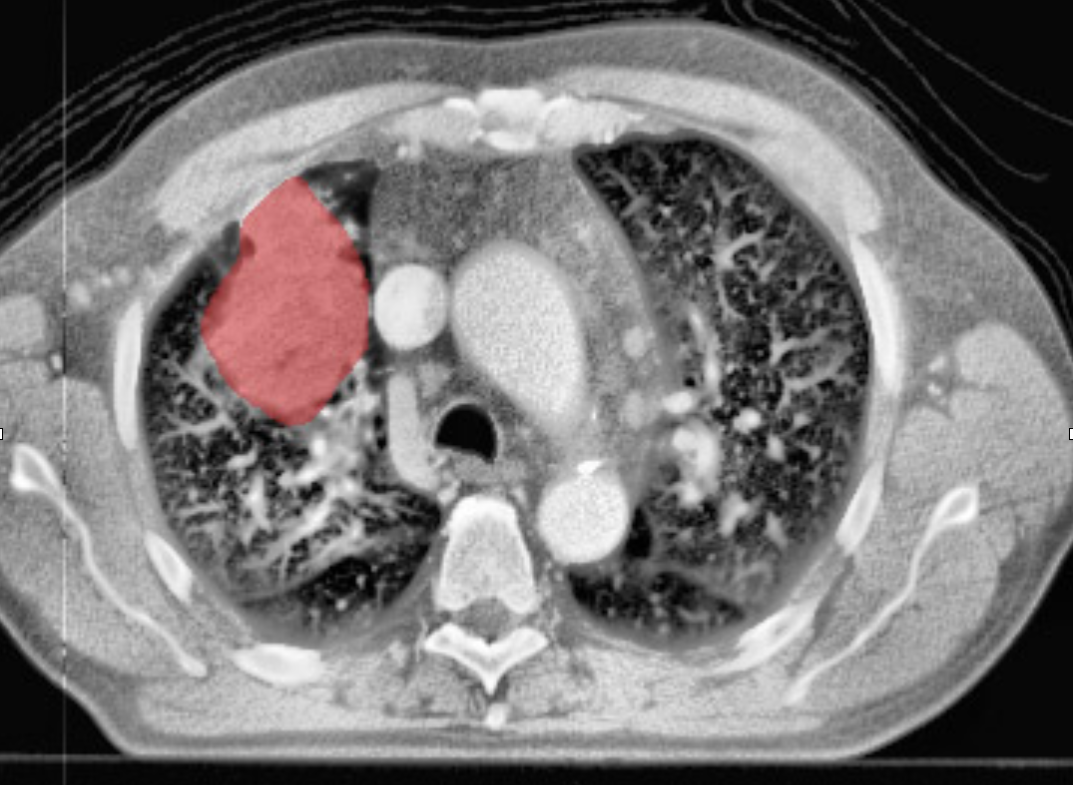}
	\caption{\small Axial slice of NSCLC patient with a survival time of 72 months. The segmented ROI is indicated in red.}
	\label{fig:image_axial_slice}
\end{figure}

\noindent The \textit{Lung1} data set is publicly available at \textit{The Cancer Imaging Archive (TCIA)}~\cite{aerts_2015} and consists of 422 NSCLC patients.
For 318 of the 422 patients, segmentations of the tumor are publicly available.
An example of an axial slice from the \textit{Lung1} dataset is provided in Fig.~\ref{fig:image_axial_slice}.

\noindent We excluded patient 176 since the segmentation mask appears to be corrupted.
Patients 72, 249, 256 and 269 were also excluded since their TNM staging indicates tumors in distant organs that may potentially corrupt a survival analysis for NSCLC.

\subsection{Baseline - Cox Proportional Hazards Model with Radiomics Features}\label{sub:Baseline - Cox Proportional Hazards Model with Radiomics Features}
\noindent As a baseline method for comparison with the proposed methods, we utilize a Cox proportional hazards model that performs hazard prediction using image features.
Based on the segmentation masks that are provided with the dataset, 18 statistics features, 15 shape features and 73 texture features based on Gray-Level-Coccurence-Matrix, Gray-Level-Runlength-Matrix, Gray-Level-Size-Zone-Matrix, Gray-Level-Difference-Matrix as well as Neighbourhood-Gray-Tone-Difference-Matrix were extracted using the PyRadiomics~\cite{griethuysen_2017} package utilizing a bin width of 25.
We deployed a forward feature selection that iteratively adds the feature with the next-highest univariate c-index to the feature set, unless its monotone Spearman correlation with a feature that is already in the feature set is higher than a threshold.





\subsection{CNN for Hazard Prediction}
\label{sub:Deep Convolutional Neural Network - Pretrained ResNet18}
\begin{figure}[h]
	\centering
	\includegraphics[width=0.3\textwidth]{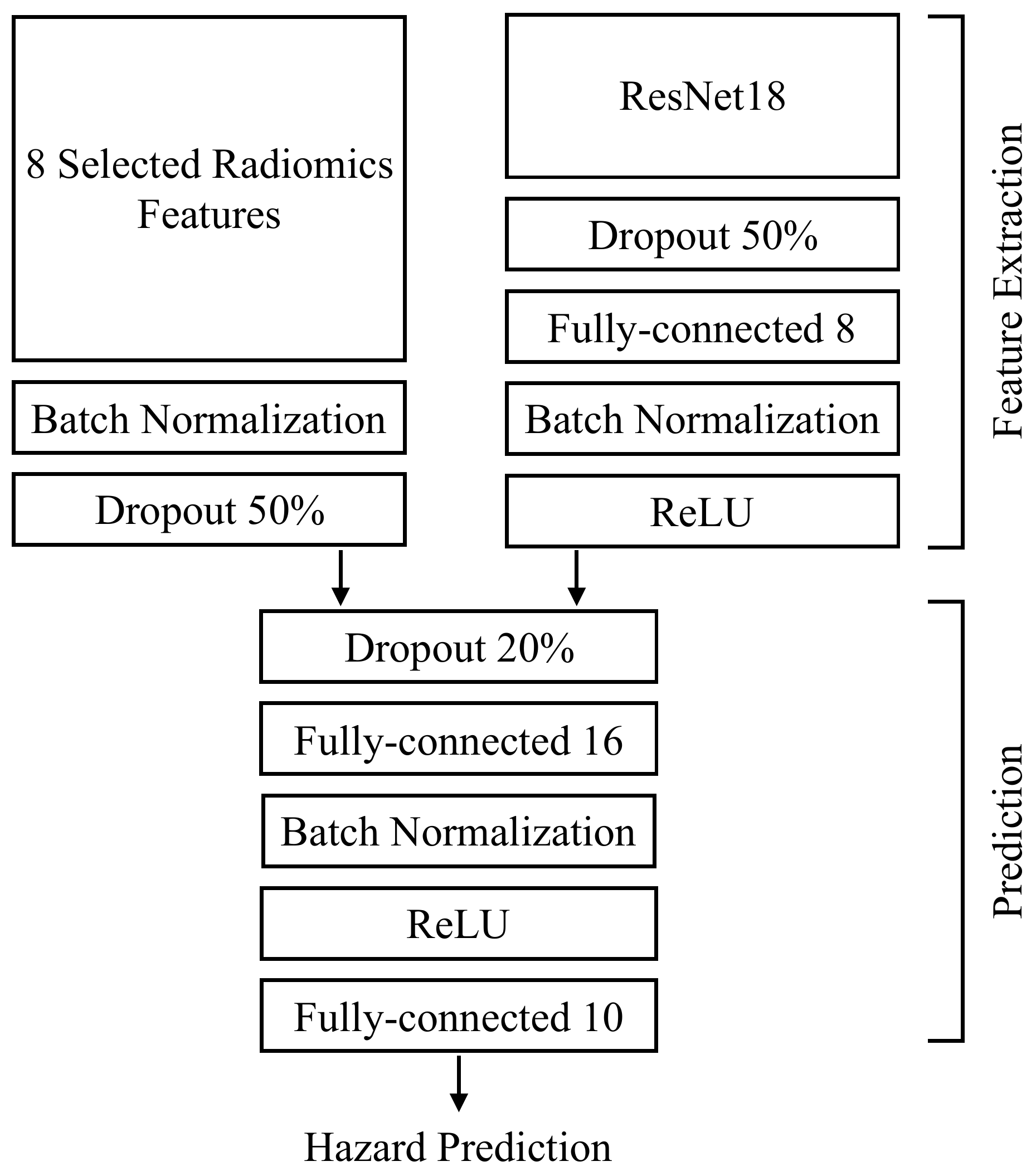}
	\caption{Model schematic for hazard prediction or classification based on both radiomics and CNN features.}
	\label{fig:dl_model}
\end{figure}

\noindent In this approach, a ResNet18~\cite{he_2016} is pretrained on the ImageNet dataset for classification of natural RGB images.
The input weights of the three RGB channels are replicated such that 25 CT slices centered around the slice containing the most tumor tissue can be utilized as input for the model.
To accommodate the entire section of the CT slices around the patient, central patches comprising \(260\times260\) pixels around the tumor centroid are extracted.
In order to adjust the ResNet18 architecture for the problem at hand, the following modifications of the architecture are performed:
The \(7\times7\) average pooling kernels are replaced by global average pooling, which makes the transition between convolutional and fully connected layers independent of the size of the consequently larger feature maps.
The CNN is used in two feature extraction approaches:
\newpage
\begin{enumerate}
	\item \textbf{CNN features:} Extract features by finetuning pretrained ResNet18 as listed in top right of Fig.~\ref{fig:dl_model}.
	\item \textbf{Multimodal features:} Concatenate radiomics features selected as explained in Section~\ref{sub:Baseline - Cox Proportional Hazards Model with Radiomics Features} with ResNet18 features. This approach is sketched by considering both blocks at top of Fig.~\ref{fig:dl_model}.
\end{enumerate}

\noindent After image feature extraction, hazard prediction is performed in two variants:
\begin{enumerate}
	\item \textbf{Direct hazard prediction:} In this setup, hazard prediction is performed by the trainable layers listed as "Prediction" in Fig.~\ref{fig:dl_model}.
	With Eqn.~\ref{eqn:CPH}, the prediction layer can be interpreted as the term \(\beta x_i\), where \(\beta\) corresponds to the weights of the layer and \(x_i\) to the activations of the previous layer instead of covariates.
	Optimizing the final fully connected layer is equivalent to the maximum likelihood estimation of \(\beta\) when fitting Cox models.
	Consequently, the resulting network can perform hazard prediction similar to~\cite{katzman_2016, zhu_2017b}.
	\item \textbf{Cox hazard prediction:} Perform hazard prediction by a Cox model and use CNN only for feature extraction after fine-tuning it with the negative partial log-likelihood.
	In this case, the "Hazard Prediction" part in Fig.~\ref{fig:dl_model} is replaced by a Cox model.
	Features are selected from the radiomics features and all activations of the fully-connected layers.
\end{enumerate}

\noindent The CNN is trained using Cox negative partial log likelihood loss
\begin{equation}
	\label{eqn:PartialLogLikelihood}
	\log L (\beta) = \sum_{T_{i,\text{uncensored}} } \beta^T x_i - \log \left(\sum_{T_j \geq t_i}\exp\left(\beta^T x_j\right) \right).
\end{equation}

\noindent However, training deep neural networks using this loss function is problematic:
Typically, training is performed using stochastic gradient descent or one of it's variants.
This works well for classification and regression problems, but when minimizing an ordering measure as in this case, the ordering problem becomes easier to solve if the batch size is small.
This is especially problematic when working with tomographic medical images and very deep network architectures because in that case, batch size \textit{must} be set very small in order to fit both model and data into GPU memory.
It is not uncommon to work with batch sizes as small as one~\cite{cicek_2016}, with which no gradient for Eqn. \ref{eqn:PartialLogLikelihood} could be computed.

\subsection{CNN for Median Survival Classification}
\label{sub:Median Survival Classification}

\noindent To overcome the shortcomings of ordering measures as loss functions it is desirable to modify the problem formulation in a way that allows training with a batch size of one.
We propose to formulate the hazard prediction problem as a classification problem.
This allows to train CNNs in a more standard setup.
The learned features can then be utilized either by a Cox model or for direct hazard prediction.
Therefore, in this approach we classify whether a patient's survival time exceeds the median survival time.
This has the additional advantage of ensuring balanced classes.
The class of a patient is then defined as 1 if the patient lived longer than the median survival time and 0 otherwise.
To incorporate censored observations for median survival classification, each patient \(i\) is assigned a weight \(w_i\) for a binary cross-entropy loss.
With the median survival time \(T_{0.5}\), the survival time \(T_i\) for patient \(i\) and the corresponding event indicator \(\delta_i\), the weights for the loss are computed according to
\begin{equation}
\label{eqn:censoring_weights}
w_{i} =
\begin{cases}
1 			& \text{if }\quad T_{0.5} \leq T_i \\
\delta_i 	& \text{if }\quad T_{0.5} > T_i
\end{cases}.
\end{equation}
This weight assumes a value of 0 for all patients censored before the median survival time and 1 otherwise.
Our complete setup looks as follows:
The CNN features from median survival classification are concatenated to the radiomics features before feature selection.
Then, a Cox proportional hazards model is fitted based on selected radiomics and CNN features to predict a hazard that allows the calculation of a c-index.

\section{Results}\label{sec:results}
\begin{table*}[]
	\begin{center}
		\begin{tabular}{l l l }
			\toprule
			\textbf{Model} & \multicolumn{2}{c}{\textbf{C-Index}} \\
			 & \textbf{Cox hazard prediction} & \textbf{Direct hazard prediction} \\
			\midrule
			Radiomics + Cox (Aerts et al.)~\cite{aerts_2014} & 0.609 $\pm$ 0.041 & - \\

			Radiomics + Cox (baseline) & 0.615 $\pm$ 0.037& - \\
			Hazard prediction CNN & \textbf{0.623 $\pm$ 0.039} & 0.585 $\pm$ 0.044 \\
			Multi-modal hazard prediction CNN & 0.620 $\pm$ 0.039 & 0.613 $\pm$ 0.04 \\
			Median survival CNN & 0.623 $\pm$ 0.04& -  \\
			Multi-modal median survival CNN   & 0.622 $\pm$ 0.038 & - \\
			\bottomrule
		\end{tabular}
	\end{center}
	\caption{
		Hazard prediction results for proposed models and baseline method.
		Reported c-indices refer to mean and standard deviations over 100 stratified random splits on the dataset.}
	\label{tab:hazard_prediction}
\end{table*}
\noindent For evaluation purposes, the data set is split into 100 random splits.
For each split, 60\,\%, 15\,\% and 25\,\% of the data is used for training, validation and testing, respectively.
The random splits are stratified based on the event indicator.
Despite the higher amount of required model fits, 100 random splits appear to be a more reliable approach to assessing the predictive accuracy of survival models than cross-validation.
While cross-validation is an appropriate evaluation method for classification or segmentation tasks, the c-index is a relative measure for predictive accuracy between individual hazard predictions.
Increasing the number of combinations of patients in different test sets therefore allows for more meaningful interpretations of the c-indices achieved.
For a fair comparison with the previous state-of-the art, we evaluate the approach from~\cite{aerts_2014}, a linear combination of four specific radiomics features in a Cox model, on the exact same data.

Tab.~\ref{tab:hazard_prediction} lists the results for the different models proposed.
The highest c-index achieved is 0.623 for a Cox model fitted with a selection of deep features and radiomics features from a hazard prediction CNN without concatenated radiomics features within the neural network.
This score is virtually identical to the corresponding c-index of 0.623 of a Cox model fitted with deep features from a median survival classification CNN and radiomics features.
Median survival time was determined using a Kaplan-Meier estimator.
The corresponding c-indices for Cox models with deep features from a multi-modal hazard prediction network and median survival classification network are lower with 0.62 and 0.622, respectively.
Direct hazard predictions from a neural network with radiomics features (multi-modal) and without are less precise with c-indices of 0.613 and 0.585 respectively.

\section{Discussion}\label{sec:discussion}

While data for 422 patients could be used for training in~\cite{aerts_2014}, our study relies on a training set containing 232 patients only.
For a fair direct comparison, we evaluated the approach from~\cite{aerts_2014} on the \textit{Lung1} dataset as it is available publicy.
Both the baseline model as well as the CNN models outperform the Cox model presented by Aerts et al.~\cite{aerts_2014}.

Relatively high variance can be observed for all models, even the linear and deterministic Cox model from~\cite{aerts_2014} with four features.
This indicates that the variance is not due to the models but rather to different generalization properties inherent to different random splits.
This is also an indication that the number of samples in the data set is insufficient.
Furthermore, it prevents a meaningful assessment of the statistical significance of the differences in c-index e.g. with a Kolmogorow-Smirnow test.
\cite{yao_2017} report similarly high variances for their experiments with pathology images.

Cox models with deep features and radiomics features outperform CNNs with concatenated radiomics features, which can be attributed to several causes.
First, overfitting might impact the CNN stronger than the Cox model despite the high drop-out for the radiomics features since there are significantly more parameters even after the concatenation than in the corresponding Cox model.
While this can be expected to influence generalization, an even more compelling second reason might be that the Cox models are fitted with 25\,\% more data relatively to the training set for the CNN models.
This is because the fine tuning of hazard CNNs proved to be highly volatile, such that a validation set is always required for early stopping, not only for hyperparameter optimization.
Another limitation of the CNN-based methods is the limited interpretability. While the model from~\cite{aerts_2014} consists of only four mathematically-defined features and a Cox model, which is straightforward to interpret, CNNs are much harder to interpret.
Nevertheless, the simplifications due to reformulation as a classification problem show promising results and could increase the accessibility of survival analysis to deep learning techniques, especially if the results can be repeated on larger datasets.
Furthermore, the technique opens up survival analysis to high dimensional image data such as 3D+t MRI or CT images that require small batch sizes when combined with CNNs.

\section{Conclusion}\label{sec:conclusion}
\noindent We presented a method for survival prediction based on tomographic medical images.
Our method can leverage trainable CNN features from CT image data, capturing abstract image information as well as clinical features in a single model.
We show that by simplifying survival analysis to median survival classification, CNNs can be trained with small batch sizes and learn features that predict survival equally well as end-to-end hazard prediction networks and outperform the previous radiomics approach.
This is a crucial step towards large scale image-based survival analysis that will allow survival prediction for more complex image data such as 3D+t images in the future.

\vfill
\pagebreak

\bibliographystyle{ieeetr}
\bibliography{/Users/christoph/Documents/LFB/Literatur/literature}

\end{document}